\pdfoutput=1

\documentclass[11pt]{article}
\usepackage[]{authblk}
\usepackage[]{acl}

\usepackage{times}
\usepackage{latexsym}

\usepackage[T1]{fontenc}

\usepackage[utf8]{inputenc}

\usepackage{microtype}

\usepackage{microtype}
\usepackage{times}
\usepackage{latexsym}
\usepackage{microtype}
\usepackage{graphicx}
\usepackage{multirow}
\usepackage{soul}
\usepackage{amsmath}
\usepackage{bbm}
\usepackage{amssymb}
\usepackage{pifont}
\DeclareMathOperator*{\E}{\mathbb{E}}
\usepackage{enumitem}
\usepackage[normalem]{ulem}
\useunder{\uline}{\ul}{}
\usepackage{setspace}
\usepackage[belowskip=0pt,aboveskip=4pt]{caption}

\usepackage{tablefootnote}
\usepackage{arydshln}
\usepackage{xspace}

\usepackage{ascii}
\usepackage{newunicodechar}
\newunicodechar{♠}{\ACK}

\usepackage{tikz}
\newcommand*\circled[1]{\tikz[baseline=(char.base)]{\node[shape=circle,draw,inner sep=0.5pt] (char) {\small#1};}}
            
\usepackage{algorithm}
\usepackage{algpseudocode}
\usepackage{csquotes}

\usepackage{etoolbox}
\usepackage{tikz}
\usetikzlibrary{tikzmark}
\usetikzlibrary{calc}

\errorcontextlines\maxdimen

\newcommand{\ALGtikzmarkcolor}{black}
\newcommand{\ALGtikzmarkextraindent}{4pt}
\newcommand{\ALGtikzmarkverticaloffsetstart}{-.5ex}
\newcommand{\ALGtikzmarkverticaloffsetend}{-.5ex}
\makeatletter
\newcounter{ALG@tikzmark@tempcnta}

\newcommand\ALG@tikzmark@start{%
    \global\let\ALG@tikzmark@last\ALG@tikzmark@starttext%
    \expandafter\edef\csname ALG@tikzmark@\theALG@nested\endcsname{\theALG@tikzmark@tempcnta}%
    \tikzmark{ALG@tikzmark@start@\csname ALG@tikzmark@\theALG@nested\endcsname}%
    \addtocounter{ALG@tikzmark@tempcnta}{1}%
}

\def\ALG@tikzmark@starttext{start}
\newcommand\ALG@tikzmark@end{%
    \ifx\ALG@tikzmark@last\ALG@tikzmark@starttext
    \else
        \tikzmark{ALG@tikzmark@end@\csname ALG@tikzmark@\theALG@nested\endcsname}%
        \tikz[overlay,remember picture] \draw[\ALGtikzmarkcolor] let \p{S}=($(pic cs:ALG@tikzmark@start@\csname ALG@tikzmark@\theALG@nested\endcsname)+(\ALGtikzmarkextraindent,\ALGtikzmarkverticaloffsetstart)$), \p{E}=($(pic cs:ALG@tikzmark@end@\csname ALG@tikzmark@\theALG@nested\endcsname)+(\ALGtikzmarkextraindent,\ALGtikzmarkverticaloffsetend)$) in (\x{S},\y{S})--(\x{S},\y{E});%
    \fi
    \gdef\ALG@tikzmark@last{end}%
}

\apptocmd{\ALG@beginblock}{\ALG@tikzmark@start}{}{\errmessage{failed to patch}}
\pretocmd{\ALG@endblock}{\ALG@tikzmark@end}{}{\errmessage{failed to patch}}
\makeatother

\usepackage{amsmath}

\algnewcommand{\LeftComment}[1]{\State \(\triangleright\) \textcolor{gray}{\textit{#1}}}
\algnewcommand{\LeftCommentNoNum}[1]{\Statex \(\triangleright\) \textcolor{gray}{\textit{#1}}}

\setlist[itemize]{leftmargin=*}

\newcommand{\nop}[1]{}
\newcommand{\hs}[1]{\textcolor{blue}{#1}}
\newcommand{\add}[1]{\textcolor{red}{#1}}

\title{
Synthetic Question Value Estimation \\ for Domain Adaptation of Question Answering
}

\author[1]{Xiang Yue}
\author[2]{Ziyu Yao}
\author[1]{Huan Sun}
\affil[1]{The Ohio State University}
\affil[2]{George Mason University}
\affil[ ]{\{\texttt{yue.149, sun.397\}@osu.edu} \; \texttt{ziyuyao@gmu.edu}}

\begin{document}
\maketitle

\begin{abstract}
Synthesizing QA pairs with a question generator (QG) on the target domain has become a popular approach for domain adaptation of question answering (QA) models. Since synthetic questions are often noisy in practice, existing work adapts scores from a pretrained QA (or QG) model as criteria to select high-quality questions. However, these scores do not directly serve the ultimate goal of improving QA performance on the target domain. In this paper, we introduce a novel idea of training a \emph{question value estimator (QVE)} that directly estimates the usefulness of synthetic questions for improving the target-domain QA performance. By conducting comprehensive experiments, we show that the synthetic questions selected by QVE can help achieve better target-domain QA performance, in comparison with existing techniques.
We additionally show that by using such questions and only around $15\%$ of the human annotations on the target domain, we can achieve comparable performance to the fully-supervised baselines.\footnote{Our source code is available at: \url{https://github.com/xiangyue9607/QVE}} 

\end{abstract}

\section{Introduction}
Question answering (QA) systems based on pretrained language models such as BERT \cite{Devlin19Bert} have recently achieved promising performance in machine reading comprehension.
However, neural QA systems trained on one domain may not generalize well to another, leaving it challenging to deploy such systems on new domains that lack large-scale QA training data\footnote{Large-scale training data are typically 60-100K in size.}. In this paper, we are interested in \emph{semi-supervised domain adaptation}: we aim to build a target QA model with source-domain data and a small number of target-domain annotated QA pairs.

\begin{figure}[t]
    \centering
    \includegraphics[width=\linewidth]{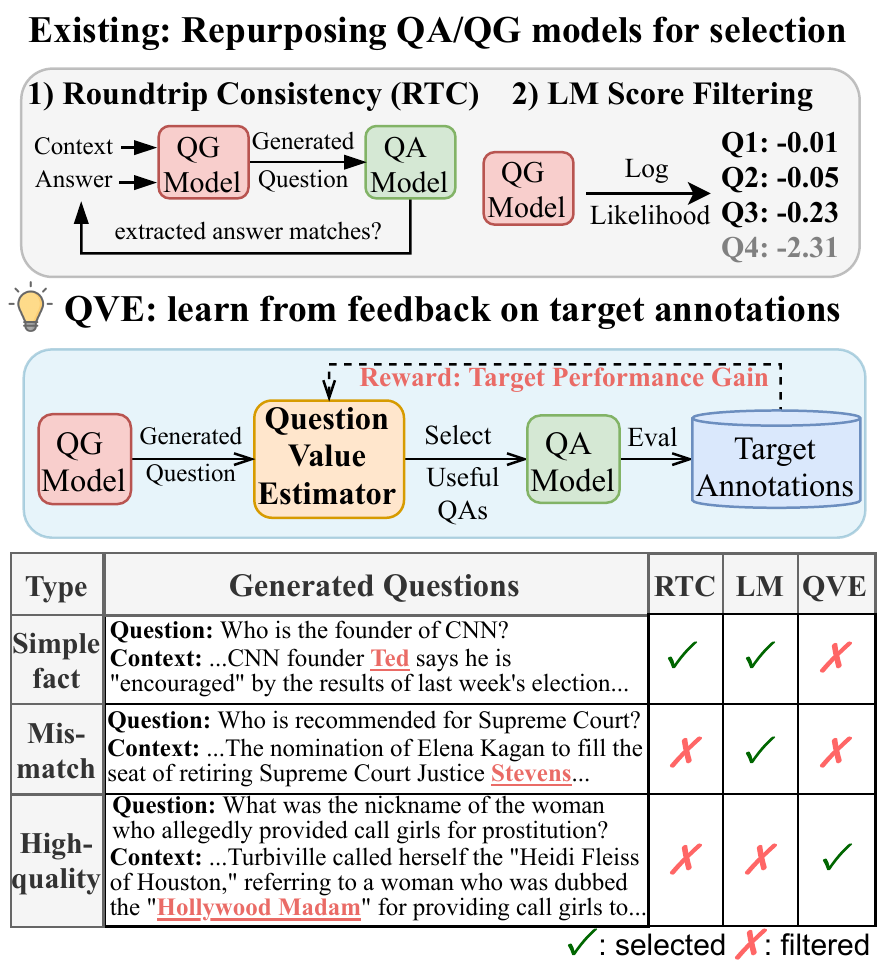}
    \vspace{-10pt}
    \caption{Existing work repurposes a pretrained QA (or QG) model to evaluate the quality of the generated questions, which is not directly associated with the target-domain QA performance and may select questions that are semantically-mismatched or ask {about} a simple fact. In contrast, our Question Value Estimator (QVE) learns to select useful questions with target-domain QA performance gain as direct feedback.}
    \vspace{-15pt}
    \label{fig:intro}
\end{figure}

Due to high annotation costs, existing work \cite{GolubHHD17,li2019Unified,WangGLLGW19,PuriSSPC20Training,chen20improved,yue2021contrastive} proposes to synthesize target-domain QA pairs via neural question generation (QG) models. The synthetic data are then used to train a QA model on the target domain. 
In practice, however, the generated questions are often of low quality, such as being semantically mismatched with their paired answers or asking {about} simple facts (Figure~\ref{fig:intro}). 
Including all such questions for QA training is less likely to bring substantial improvements. This inspires us to study a crucial problem:

\vspace{1pt}
\textit{Given a set of target-domain synthetic QA pairs, how to select high-quality ones that are useful to improve target-domain QA training?}
\vspace{1pt}

To address the problem, \citet{Alberti19Synthetic} propose the Roundtrip Consistency (RTC) method, {which filters\footnote{We interchangeably use ``filter'' (noisy/low-quality questions) and ``select'' (useful/high-quality questions).} questions that cannot be correctly answered by a pretrained QA model}. Other work \cite{ShakeriSZNNWNX20} considers using the generation log likelihood by the QG model (LM Score) as a metric to filter noisy questions (Figure~\ref{fig:intro}, top).
Although these filtering techniques have been shown to improve the question quality to some extent \cite{RennieMMNG20}, they are not directly optimized for \emph{selecting questions that can improve QA performance on the target domain}. For example, some useful but difficult questions (e.g., the last example in Figure~\ref{fig:intro}) may be filtered by the Roundtrip method, since they cannot be answered correctly by the pretrained QA model. However, these questions are often crucial to further improving QA performance when added into training.

In this paper, we propose a \emph{question value estimator (QVE)} (Figure~\ref{fig:intro}, middle) to select questions that can improve QA performance on the target domain. QVE takes in generated QA examples and outputs real-valued scores (i.e., question values), which are expected to represent the usefulness of generated questions in terms of improving target-domain QA performance. However, training the QVE model towards this goal is challenging due to the lack of supervision (i.e., true question values). 


To solve the problem, we propose to train the QVE with direct QA feedback from the target domain. Intuitively, \nop{if a batch of synthetic questions (used for training QA) leads to an accuracy increase of the target-domain QA model, QVE should assign high values to such questions. The larger the accuracy increase, the higher the question values they should be assigned.}if a batch of synthetic questions (when used for training) leads to increasing accuracy of the target-domain QA model, QVE should assign high values to them; the more the accuracy increases, the higher the question values should be.  Thus, we optimize QVE with the \textit{target-domain QA performance gain} after adding the selected questions into training. More formally, given the discrete and non-differentiable question selection process, we formulate the question selection of QVE as a reinforcement learning~\cite{Williams92} problem (Figure \ref{fig:model}). The QVE receives a batch of synthetic samples each time and learns to select high-quality ones based on their estimated values. The selected samples are then used to train the target-domain QA model, with the resulting performance gain (on the available target-domain annotations) as the reward. The reward guides the optimization of QVE such that it will eventually make proper question value estimation and selection.

{To evaluate the QVE model, we instantiate the QG and the QA model based on the pretrained BART~\cite{Lewis20BART} and BERT~\cite{Devlin19Bert}, respectively.} By carrying out comprehensive experiments on four commonly-used reading comprehension datasets \cite{TrischlerWYHSBS17,JoshiCWZ17,Yang0ZBCSM18,KwiatkowskiPRCP19}, we show that: 
(1) our QVE model trained with the target-domain QA feedback substantially outperforms the question selection techniques trained without direct QA feedback \cite{Alberti19Synthetic, ShakeriSZNNWNX20}. 
(2) When using our QVE model to select synthetic questions, QA models can achieve comparable performance to fully-supervised baselines while using only 15\% of the full target-domain annotations, which indicates that our method can greatly alleviate human annotation effort in practice.
(3) To understand why QVE brings superior improvement, we conduct human evaluation and find that QVE can better identify semantically-matched and difficult questions. 

\section{Related Work}
\noindent\textbf{Domain Adaptation of Question Answering.}
In this field, some work \citep{wiese2017neural, chung2018supervised, hazen2019towards, cao2020unsupervised} assumes that target-domain annotated questions are available, however, manually creating questions is costly. Therefore, another line of research work \cite{GolubHHD17,WangGLLGW19,LeeLJKH20,ShakeriSZNNWNX20} investigates a domain adaptation setting where annotated questions are not available on the target domain. A commonly-adopted approach of this line is to leverage a neural question generation (QG) model \citep{du2017learning, zhou2017neural, sun2018answer, zhao2018paragraph, nema2019let, tuan2020capturing} to automatically synthesize questions given unlabeled contexts \cite{du2018harvesting, ZhangB19Addressing,WangGLLGW19,liu2020asking, GolubHHD17,WangGLLGW19,LeeLJKH20,ShakeriSZNNWNX20,yue2021contrastive}; see more discussions in Section \ref{sec:background}. However, it is very challenging to achieve satisfying performance without any target annotations. In our work, we study \textit{semi-supervised domain adaptation of QA}, and assume \textit{a small number of target annotations are available}, which can greatly help models adapt to the target domain while requiring minimal human effort.

\noindent\textbf{Unsupervised and Semi-supervised QA} are two other research topics relevant to our work \cite{fabbri2020template,li2020harvesting,LewisDR19,dhingra2018simple}. Unlike domain adaptation, these two settings do not assume the existence of the ``source domain'' and synthesize cloze-style questions via rule-based methods for building QA models. Since rule-based QG methods typically have much worse performance than neural ones (pretrained on the source data), we do not compare with these two lines of research in experiments.

\noindent\textbf{Data Selection} methods aim to select a useful subset from the (noisy) training data. Though (RL-based) data selection methods were explored in other NLP tasks~\cite{emnlp/RuderP17,wsdm/QuJQYMCHC19,acl/LiuSZZ19}, none of them can be directly applied with trivial efforts to our QA scenario and semi-supervised setting. For example, \cite{emnlp/RuderP17} and \cite{acl/LiuSZZ19} reward or measure the selection with the distribution distance between the selected data and target data, while we reward the selection by measuring how large the improvement the selected data can bring for target-domain QA training, which is more aligned with the end goal. Our work is mostly inspired by recent research on data selection in machine learning community \cite{GhorbaniZ19,JiaDWHHGLZSS19}, particularly \cite{yoon2020data}.  However, the significant differences between our work and \cite{yoon2020data} are as follows: 1) we study a very challenging task, domain adaptation of question answering, which was not studied in \cite{yoon2020data}. How to develop a method in a similar spirit for this task is unexplored. 2) In order to study the task, we begin our method by first proposing two data selection methods that are not covered in \cite{yoon2020data} but achieve comparable results to existing baselines. We then introduce our RL-based method with a carefully-designed reward, which is well connected to the end goal of improving target-QA performance.

\section{Background}
\label{sec:background}

\subsection{Domain Adaptation of QA via QG}
\noindent\textbf{Semi-supervised Domain Adaptation.}
We study the semi-supervised domain adaptation of \textit{extractive} question answering, where the source-domain and a small number\footnote{In our  experiments, we assume 1,000 target annotations available, which is around 1-1.5\% of the original training data.} of target-domain QA annotations are provided.
Formally, we denote the source-domain QA dataset as $D^\mathsf{s}=\{(c_i^\mathsf{s}, q_i^\mathsf{s},a_i^\mathsf{s})\}_{i=1}^N$, where large-scale tuples of context $c_i^\mathsf{s}$, question $q_i^\mathsf{s}$, and answer $a_i^\mathsf{s}$ are available. For the target domain, only a small set of annotated QA pairs $D^{\mathsf{t}}=\{(c_j^\mathsf{t}, q_j^\mathsf{t},a_j^\mathsf{t})\}_{j=1}^M$ are available ($M \ll N$).  Since unlabeled contexts are easy to collect, we assume that they are largely available: $C^\mathsf{t}=\{c_l^\mathsf{t}\}_{l=1}^L$ ($L\gg M$). The task is to build a QA model that can accurately answer questions on the target domain, given $D^\mathsf{s}$, $D^{\mathsf{t}}$, and $C^\mathsf{t}$.\nop{to train (or finetune) a QA model $f_\theta$ that can \nop{be used to}accurately answer questions on the target domain, given $D^\mathsf{s}$, $D^{\mathsf{t}, v}$, and $C^\mathsf{t}$.}

\noindent\textbf{Domain Adaptation via Question Generation.} Given the lack of large-scale target-domain annotations, an intuitive approach to domain adaptation is first synthesizing target-domain QA data ${D}^\mathsf{t}_{syn} = \{(c_l^\mathsf{t}, q_l^\mathsf{t}, a_l^\mathsf{t})\}_{l=1}^L$ automatically from the unlabeled contexts $C^\mathsf{t}$, and then training a target-domain QA model on the synthetic (${D}^\mathsf{t}_{syn}$) and the small-size annotated ($D^{\mathsf{t}}$) target-domain data. In such an approach, a question generator (QG) $g_\phi$ is first pretrained on the source training data and further finetuned on the available target-domain annotated QA pairs. A well-trained QG model then takes target-domain context-answer pairs as input to generate a question: $q_l^\mathsf{t}=g_\phi (c_l^\mathsf{t}, a_l^\mathsf{t})$. \nop{With respect to how to obtain $a_l^\mathsf{t}$, we assume an answer $a_l^\mathsf{t}$ (i.e., a text span on the context $c_l^\mathsf{t}$) is given, following \citet{du2017learning}. When the answer $a_l^\mathsf{t}$ is not given, it can be specified in\st{extracted from} the given context by using an entity recognition tool \cite{du2018harvesting}, a classifier \cite{PuriSSPC20Training} or a seq2seq model \cite{ShakeriSZNNWNX20, yue2021contrastive}. \hs{In this work, we will mainly focus on selecting useful synthetic questions}\nop{As we mainly focus on\st{estimating question values and} selecting useful \hs{synthetic} questions in this work, \add{we leave the exploration of answer extraction as future work.}}
}

Although this approach has been shown promising, in practice, its effectiveness is restricted by the quality of synthetic questions. Thus, learning to select ones that can lead to a better target-domain QA model becomes a crucial problem. 

With respect to how to obtain $a_l^\mathsf{t}$ for QG, in this paper, we assume an answer $a_l^\mathsf{t}$ (i.e., a text span in the context $c_l^\mathsf{t}$) is given, following \citet{du2017learning}. When the answer $a_l^\mathsf{t}$ is not given, it can be extracted from the given context by using an entity recognition tool \cite{du2018harvesting}, a classifier \cite{PuriSSPC20Training} or a seq2seq model \cite{ShakeriSZNNWNX20}. Note that noise caused by such answer extraction tools will further lower the overall quality of the synthesized questions. In this paper, we focus on how to select useful synthetic questions in general (i.e., those questions can be synthesized by any QG process) and assume answers are given for simplicity.

\subsection{Synthetic Question Selection} 
Given the synthetic target-domain QA data ${D}^\mathsf{t}_{syn}$, the task is to select high-quality pairs from ${D}^\mathsf{t}_{syn}$ that are useful to improve target-domain QA training. Such a selection decision is often made based on some scores that can indicate the quality of the pairs. For example, Roundtrip filtering \cite{Alberti19Synthetic} selects questions based on the extracted answer's correctness by a pretrained QA model. Similarly, LM filtering  \cite{ShakeriSZNNWNX20} selects questions with high log-likelihood scores in the generation. 
However, these scores do not directly serve the goal of improving target-domain QA training. Inspired by recent research on data selection in the machine learning community \cite{GhorbaniZ19,JiaDWHHGLZSS19,yoon2020data}, we propose a new idea of training a \textit{question value estimator}, which predicts the usefulness of a synthetic question for target-domain QA. 
\begin{figure*}[t]
    \centering
    \includegraphics[width=0.9\linewidth]{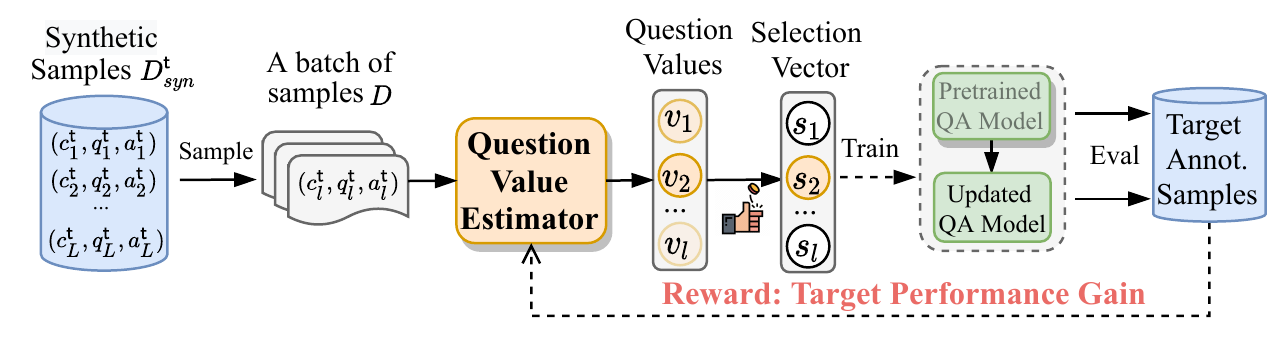}
    \caption{Illustration of QVE training based on the direct feedback from QA. Specifically, in the forward pass, QVE estimates the question values of a batch of synthetic questions and draws a Bernoulli sampling to select questions. The selected questions are then used to finetune a pretrained QA model. The performance gain (before and after the QA finetuning) on the target annotations is calculated as the reward for REINFORCED QVE training. }
    \label{fig:model}
\end{figure*}
\section{Question Value Estimator (QVE)}
Formally, we design a question value estimator (QVE), $e_\gamma$,  which takes in a synthetic QA example $(c_l, q_l, a_l)$ (for simplicity, we omit the superscript $\mathsf{t}$) and outputs a score indicating its ``value,'' i.e., $v_l=e_\gamma(c_l, q_l, a_l)$. The ``value'' can imply ``the potential for improving the target-domain QA performance when being used as a training sample''. With this score, one can select most useful synthetic examples for the target-domain QA training. 

We use a BERT model as the backbone of the QVE. Specifically, we concatenate the context, question and answer as input to the QVE, and use BERT to encode the sequence \cite{Devlin19Bert}. 
\begin{equation*}
    \mathbf{h} = \texttt{BERT}\;[\texttt{<CLS>}\; q\;\texttt{<ANS>\;}a\;\texttt{<SEP>}\;c]
\end{equation*}
where $q,a,c$ represent the question, answer, and context, respectively. $\mathbf{h} \in \mathbb{R}^H$ denotes the hidden representation of the input sequence derived from the ``\texttt{<CLS>}'' token. \texttt{<ANS>} and \texttt{<SEP>} are two special tokens used as delimiters.

In our preliminary experiments, we find that adding the answer (start index and end index) probabilities ($p_s,p_e$) by a pretrained QA model as additional features to the hidden representation $\mathbf{h}$ can accelerate the QVE training convergence and lead to better performance. Thus, we add these two features ($p_s,p_e$) followed by linear transformations of the original hidden representation, and then build a linear classifier to output the question value.
\begin{gather*}
    \mathbf{h'}=\sigma(W_2\sigma(W_1 \mathbf{h}+b_1)+b_2)\\
    \mathbf{h''}=\sigma(W_3(\mathbf{h'}\oplus p_s \oplus p_e)+b_3)\\
    v_l = W_4 \mathbf{h''} + b_4
\end{gather*}
where $W_1\in \mathbb{R}^{H_1 \times H}, W_2\in \mathbb{R}^{H_2 \times H_1}, W_3\in \mathbb{R}^{H_3 \times H_2}, W_4\in \mathbb{R}^{H_3}$, $b_1\in \mathbb{R}^{H_1}, b_2\in \mathbb{R}^{H_2}, b_3\in \mathbb{R}^{H_3}, b_4\in \mathbb{R}$ are trainable parameters of linear layers. $\sigma$ is the activation function \texttt{tanh}.

Learning such a question value estimator is challenging because we do not have direct supervision on the true value or usefulness of a synthetic question. We discuss two straightforward baselines to train QVE in Section \ref{sec:qve_train_baseline}, and a more advanced one based on reinforcement learning in Section \ref{sec:qve_train_rl}.

\subsection{QVE Training: Two Baselines}
\label{sec:qve_train_baseline}

\noindent\textbf{Binary Classifier}: One straightforward solution is to treat QVE as a binary classifier and train it based on the human-annotated (positive) and the machine-synthesized (negative) QA pairs. Given the scarcity of target-domain data, we first pretrain the classifier on the source domain and then finetune it on the target domain.
More specifically, we train a QG model on 70\% of the source training data and generate synthetic questions on the remaining 30\% of the source training contexts. The generated questions and the source-domain annotated questions are used to train this binary classifier. The classifier is then finetuned based on the small set of target-domain annotations (positive) and the  samples synthesized on the same target-domain contexts (negative).

However, not all of the generated questions are bad. Simply treating all synthetic samples as negatives may mislead the classifier. Thus, we loose this assumption and introduce a ranking baseline.

\noindent\textbf{Ranking Baseline}: We assume that the quality of human-annotated questions is not inferior than that of machine-synthesized ones. Thus, we train QVE based on a ranking triplet loss defined as follows:
\begin{equation*}
    L_{r}=\sum \max(0,m+v_s-v_h)
\end{equation*}
where $v_s, v_h$ are the estimated question values of the machine-synthesized sample and human-annotated sample. $m$ is set to $0.15$ as the margin.


The two baseline methods have two obvious drawbacks: (1) they are trained to differentiate between human-annotated and machine-synthesized samples, which is mismatched with our goal of selecting high-quality samples \emph{among machine-synthesized data}; (2) similar as \cite{Alberti19Synthetic, ShakeriSZNNWNX20}, the two baselines are \emph{not} trained with direct signals that can represent the usefulness of a synthetic question. In the next section, we will introduce a task-specific training method, which directly uses the target-domain QA feedback to optimize QVE.


\setlength{\textfloatsep}{8pt}

\begin{algorithm}[!t]
 \caption{QVE REINFORCED Training }
 \label{alg:model_qve}
 \textbf{Input}: pretrained QA model $f_\theta$; target synthetic QA pairs ${D}^\mathsf{t}_{syn}$; small target annotations $D^\mathsf{t}$.\\
 \textit{Hyperparameters}: outer iterations $I_o$, outer batch size $B_o$, inner iterations $I_n$, inner batch size $B_n$, QVE learning rate $\alpha_o$, QA learning rate $\alpha_n$.
 
\textbf{Output}: QVE $e_\gamma$.

\begin{algorithmic}[1]
\State Randomly initialize $e_\gamma$
\State Store $\theta_0 \leftarrow \theta$ (pretrained QA checkpoint)
\For{$\text{outer iteration}=1$ to $I_o$}
\LeftComment{\circled{1} Sample a batch of synthetic QA pairs:}
\State  Sample $\mathcal{D} = \{(c_l, q_l, a_l)\}_{l=1}^{B_o}$ from ${D}^\mathsf{t}_{syn}$
\LeftComment{\circled{2} Estimate question values:}
\State  $\mathcal{V}=e_\gamma(\mathcal{D})$
\LeftComment{\circled{3} Sample selection vector: }
\State  $\mathcal{S} \sim \texttt{Bernoulli} (\mathcal{V})$
\LeftComment{\circled{4} Update QA on selected samples:}
\For{$\text{inner iteration}=1$ to $I_n$}
    \State Sample $\{(c_l, q_l, a_l)\}_{l=1}^{B_n} \sim \mathcal{D}$
    \State $\theta \leftarrow \theta-\frac{\alpha_n}{B_n}\sum_{l=1}^{B_n} s_{l} \cdot \nabla_\theta \mathcal{L}_{qa}$ 
\EndFor

\LeftComment{\circled{5} Calculate QA gain as QVE reward:}
\label{algoline:reward_func}\State $r_{qve}= \texttt{reward\_fn} (f_{\theta_0}, f_\theta, D^{\mathsf{t}})$

\LeftComment{\circled{6} Update QVE based on Eq.~\ref{eq:nabla_gamma}:}
\State $\gamma \leftarrow \gamma-\alpha_o \cdot \nabla_\gamma \mathcal{L}_\gamma$
\State Reset $\theta \leftarrow \theta_0$
\EndFor

\State \textbf{return} $e_\gamma$
\end{algorithmic}
\end{algorithm}

\subsection{QVE Training: Direct Feedback from QA}
\label{sec:qve_train_rl}
A well-trained QVE is expected to assign high values to synthetic questions that can improve the target-domain QA performance.
Therefore, an intuitive way to measure the value of a synthetic question is to consider the downstream QA performance gain (on the available target annotations) before and after this question is included in the training set.
However, this ``leave-one-out'' formulation is computationally expensive and time-consuming, given that it can estimate the value of only one single synthetic question in each forward pass. In light of this challenge, we instead estimate question values in a \emph{batch-wise} fashion.  Algorithm~\ref{alg:model_qve} and Figure \ref{fig:model} describe the learning process.

Generally speaking, we frame the QVE model learning as a reinforcement learning problem \cite{Williams92}, and stimulate QVE to assign higher values to more useful questions by using performance-driven rewards. Specially, for a batch of synthetic examples $\mathcal{D} = \{(c_l, q_l, a_l)\}_{l=1}^{B_o}$ in the outer training iteration (Line 4-5), the QVE model \emph{selects a subset of examples} that are most likely to boost the QA performance on the target domain, based on its judgment on their values.

Mathematically, the decision-making outcome is represented by the selection vector {$\mathcal{S}=(s_1, s_2, ..., s_{B_o})$, where $s_l \in \{0, 1\}$} $l=1,...,B_o$ (Line 6-9). The whole batch-level decision making policy $\pi_\gamma$ is described as follows:
\begin{gather*}
    v_l = e_\gamma(c_l, q_l, a_l)\\
    s_l \sim \texttt{Bernoulli} (v_l) \\
    \pi_\gamma(\mathcal{S}|\mathcal{D}) = \prod_{l=1}^{B_o}[v_l^{s_l} \cdot (1-v_l)^{1-s_l}], 
\end{gather*}
where the selection of a certain example $(c_l, q_l, a_l)$ is formulated as sampling from a Bernoulli distribution of probability $v_l$ (i.e., its estimated question value). We adopt the Bernoulli sampling based on the estimated value $v_l$ instead of setting a hard threshold to encourage the policy exploration.

The model is rewarded based on how much performance gain the selected examples could bring when they are used to train the target-domain QA model. To this end, we finetune the QA model $f_\theta$ on the selected batch samples based on $\mathcal{L}_{qa}$, which typically is a cross-entropy loss:
\begin{equation*}
    \mathcal{L}_{qa} = -\sum_{l}^{B_o} \log P (a_l|q_l,c_l;\theta)
\end{equation*}
In practice, to stabilize the QVE training, we choose a large outer batch size $B_o$ in each outer training iteration. For finetuning the QA model, we pick a relatively smaller inner batch size $B_n$ and repeat the training for $I_n$ times, such that the QVE-selected samples are fully utilized (Line 10-14).

The reward $r_{qve}$ is defined as the QA performance gain on the target-domain annotations $D^{\mathsf{t}}$ before ($f_{\theta_0}$) and after ($f_\theta$) finetuning (Line 15-16), 
\begin{equation*}
    r_{qve}= \texttt{reward\_fn} (f_{\theta_0}, f_\theta, D^{\mathsf{t}})
\end{equation*}
where \texttt{reward\_fn} is Exact Match (EM) gain\footnote{We also tried F1 gain and loss drop as the \texttt{reward\_fn} and the EM gain is slightly better than the other two.}.

Given the discrete and non-differentiable question selection process, we update the QVE model using the REINFORCE algorithm \cite{Williams92}. Mathematically, we aim to minimize:
\begin{equation*}
    \mathcal{L}_\gamma = - \E_{\mathcal{S} \sim \pi_\gamma(\cdot|\mathcal{D})}[r_{qve}].
\end{equation*}
The gradient of the loss function is derived as:
\begin{equation}
\label{eq:nabla_gamma}
\begin{aligned}
    &\nabla_\gamma \mathcal{L}_\gamma
    = - \E_{\mathcal{S} \sim \pi_\gamma}[r_{qve} \nabla_\gamma \log \pi_\gamma(\mathcal{S}|\mathcal{D})]\\
    &= - \E_{\mathcal{S} \sim \pi_\gamma}[r_{qve} \nabla_\gamma \sum_{l=1}^{B_o}\log[v_l^{s_l}  (1-v_l)^{1-s_l}]].
\end{aligned}
\end{equation}
Notably, to mitigate the instability in reinforcement learning, we reset the QA model to its pretrained checkpoint at the end of each outer iteration (Line 19), and keep the pretrained QG model unchanged.

After training QVE, we can use it to calculate the question value for all the synthetic questions on the target domain. Then we can select top $K\%$ synthetic QA pairs as the training corpus to train the target-domain QA model.

\section{Experimental Setup}
\subsection{Datasets}
We use datasets in the MRQA 2019 Shared Task \cite{FischTJSCC19}, a popular challenge focusing on generalization in reading comprehension. Specifically, following \citet{ShakeriSZNNWNX20}, we use \textbf{SQuAD 1.1} \cite{RajpurkarZLL16} as the \textit{source-domain} dataset. For the \textit{target-domain} datasets, we consider \textbf{NewsQA} \cite{TrischlerWYHSBS17}, \textbf{Natural Questions (NQ)} \cite{KwiatkowskiPRCP19}, \textbf{HotpotQA} \cite{Yang0ZBCSM18} and \textbf{TriviaQA} \cite{JoshiCWZ17} as they are commonly used and have sufficient contexts for the QG model to generate synthetic samples. Since there is no test set available for each dataset, we use the original dev set as the test set. Detailed descriptions of each dataset are in Appendix \ref{apdx:datasets}. 

For the target-domain datasets, we assume all the contexts and $n$ annotated QA pairs in the original training sets are available for training. We set $n=1000$ (about 1\%-1.5\% of original training sets) as default and discuss the impact of $n$ in Section \ref{sec:target_dev_size}.

\subsection{Implementation Details}

We implement models using the Hugging Face transformers \cite{wolf-etal-2020-transformers} library. We instantiate the QA model with \texttt{BERT-base-uncased} \cite{Devlin19Bert}, and the QG model with \texttt{BART-base} \cite{Lewis20BART}. For training  QVE (Algorithm~\ref{alg:model_qve}), we use \texttt{BERT-base-uncased} model and set $H_1=H_3=H=768$ and $H_2=64$ for linear layers. To enable a large batch size $B_o$, we use gradient checkpointing \cite{corr/ChenXZG16}, a technique used for reducing the memory footprint when training deep neural networks. We set $I_o=2000$, $B_o=80$, $I_n=20$, $B_n=4$, and $\alpha_o=\alpha_n=3e^{-5}$. To select the best QVE checkpoint, we pick the one that achieves the highest reward on the target annotations or the one that leads to the lowest QA training loss. When training (finetuning) QA and QG models (either on source or target domain), we set training epochs as 2 and 3 respectively. Other hyperparameters are set as default in the transformers library. 



\subsection{Compared Baselines}
We evaluate the following QA models built on different training data:

    \noindent \textbf{(1) Source Only Baseline}: we train a QA model on the source-domain data.
    
    \noindent \textbf{(2) Source + Target Annotations Baseline}: we further finetune the ``(1) Source Only Baseline'' on the available target annotated QA pairs.
    
    \noindent \textbf{(3) QG Baseline (no filtering)}: we first pretrain a QG model on the source-domain data and finetune it on the available target annotations. The QG model is then used to generate synthetic QA samples on the target contexts. We finetune a QA model sequentially on all available data with the order of ``source$\rightarrow$target synthetic$\rightarrow$target annotated'' for all the datasets except TriviaQA\footnote{For the TriviaQA dataset, we merge the target synthetic and target annotated dataset into one training file since directly finetuning on the target annotated dataset would hurt the QA performance based on our preliminary experiments. }. The same QA finetuning strategy will also be used for (4)-(8).
    
    \begin{table}[t]
\centering
\resizebox{0.9\linewidth}{!}{%
\begin{tabular}{l|cccc}
\hline
 & \multicolumn{4}{c}{Different Filtering Methods} \\ \hline
Dataset & NoFilter & RTC & LM & QVE \\ \hline
NewsQA & 74,160 & 33,756 & 44,485 & 44,485 \\
NQ & 104,071 & 62,888 & 62,443 & 62,443 \\
HotpotQA & 72,928 & 46,273 & 43,757 & 43,757 \\
TriviaQA & 61,688 & 26,361 & 37,013 & 37,013 \\ \hline
\end{tabular}
}
\caption{Number of synthetic examples selected by different methods. NoFilter: QG baseline (no filtering); RTC: Roundtrip Filtering; LM: LM Filtering.}
\label{apdx_tbl:number_synthetic_examples}
\end{table}

    \noindent \textbf{(4) RoundTrip Filtering} \cite{Alberti19Synthetic}: we use the ``(2) Source + Target Annotation Baseline'' to extract answers for target synthetic questions and select the ones, whose extracted answers are correct, as the target synthetic training corpus.
    
    \noindent \textbf{(5) LM Filtering} \cite{ShakeriSZNNWNX20}:
    we use the log likelihood scores of synthetic questions produced by the QG model in (3) as the filtering criterion. We select top K\% samples as the target synthetic training corpus.
    
    \noindent \textbf{(6) QVE (binary classifier)}: we train QVE as a binary classifier (Section \ref{sec:qve_train_baseline}) and then use it to select top K\% target synthetic samples.
    
    \noindent \textbf{(7) QVE (ranking baseline)}: we train QVE based on a ranking function (Section \ref{sec:qve_train_baseline}), and then use it to select top K\% synthetic samples.
    
    \noindent \textbf{(8) QVE (RL)}: we train QVE based on the direct feedback from target annotations using RL (Section \ref{sec:qve_train_rl}), and then use it to select top K\% target synthetic samples. 
    
    \noindent \textbf{(9) Fully-supervised Baseline}: we train a QA model on the original target training data. Note that we report the fully-supervised performance here only as the reference and (1)-(8) are not directly comparable to this.   
    
\begin{table*}[t]
\centering
\resizebox{0.9\linewidth}{!}{%
\begin{tabular}{l|l|cc|cc|cc|cc}
\hline
\multicolumn{1}{l|}{\multirow{2}{*}{\begin{tabular}[c]{@{}c@{}}No.\end{tabular}}}  & \multirow{2}{*}{Methods} & \multicolumn{2}{c|}{NewsQA} & \multicolumn{2}{c|}{NQ} & \multicolumn{2}{c|}{HotpotQA} & \multicolumn{2}{c}{TriviaQA} \\\cline{3-10} 
 &  &  EM & F1 & EM & F1 & EM & F1 & EM & F1 \\ \hline
(1)  & Source Only Baseline & 40.2 & 56.2 & 45.2 & 59.1 & 43.3 & 60.3 & 49.5 & 59.3 \\ \hline
(2)   & Source + Target Annotations Baseline & 43.7 & 59.8 & 54.2 & 68.2 & 51.7 & 69.2 & 55.7 & 62.0 \\ \hline

(3) & QG Baseline (no filtering) & 45.3 & 60.7 & 60.5 & 72.6 & 52.9 & 70.0 & 58.3 & 63.9 \\
(4)   & \,+RoundTrip Filtering \cite{Alberti19Synthetic} & 45.4 & 60.8 & 58.6 & 71.2 & 53.9 & 70.5 & 58.7 & 64.4 \\
(5)   & \,+LM Filtering \cite{ShakeriSZNNWNX20}
 & 45.3 & 61.2 & 60.0 & 72.1 & 53.9 & 70.5& 56.0 & 61.7 \\
(6)   & \begin{tabular}[c]{@{}c@{}} \,+QVE (binary classifier)\\\end{tabular} 
 & 45.2 & 60.7 & 60.1 & 72.3 & 53.7 & 70.4 & 58.2 & 63.8 \\
(7)   & \begin{tabular}[c]{@{}c@{}} \,+QVE (ranking baseline) \\\end{tabular} & 45.8 & 61.3 & 60.6 & 72.8 & 53.9 & 70.9 & 58.4 & 63.9 \\
(8)   & \begin{tabular}[c]{@{}c@{}}\textbf{ +QVE (RL)} \\\end{tabular}
  & \textbf{46.2} & \textbf{61.6} & \textbf{61.3} & \textbf{73.2} & \textbf{54.5} & \textbf{71.7} & \textbf{62.3} & \textbf{68.5} \\ \hline
(9)   & Fully-supervised Baseline & 50.0 & 64.6 & 65.8 & 78.1 & 56.8 & 73.9 & 64.6 & 70.3 \\ \hline
\end{tabular}%
}
\caption{Semi-supervised domain adaptation performance of different models where 1,000 target-domain annotations (around 1-1.5\% of the original training data) are used. }
\label{tbl:overall_performance}
\end{table*}
\begin{figure*}[t]
    \centering
    \includegraphics[width=\linewidth]{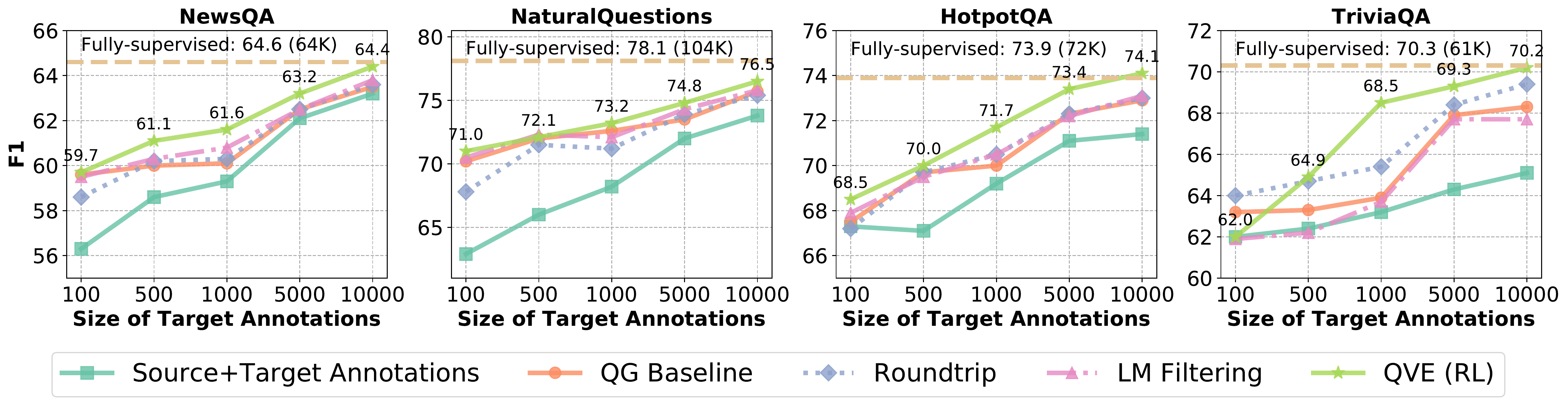}
    \caption{Impact of the number of target annotated QA pairs. We also show the fully-supervised performance (and \#train) as the reference. With 10K target annotations (around 15\% of the full training set), our method can achieve comparable performance to the supervised ones (as shown at the top of each sub-figure).}
    \label{fig:number_of_target_samples}
\end{figure*}

The number of the selected synthetic examples of RoundTrip Filtering is determined by the QA model and varies for each dataset. For LM Filtering and QVE, we select top K\% (K=60) samples among all synthetic ones and discuss the impact of the synthetic dataset size in Appendix \ref{sec:impact_syth_datasize}. We show the statistics of filtered datasets in Table \ref{apdx_tbl:number_synthetic_examples}.

\section{Results}

\subsection{Overall Results}
We first discuss the domain adaptation results on the 4 target-domain QA datasets under semi-supervised setting where $n=1,000$ target-domain QA examples are available. Table \ref{tbl:overall_performance} shows the overall results of different methods. We summarize key findings as follows:

\noindent (1) Compared with RoundTrip and LM Filtering, our QVE (RL) achieves the best performance\nop{ under the three training settings}. This is because both baselines are not specifically trained to select useful examples for improving QA performance on the target domain. Our QVE, on the contrary, is trained with a signal that directly reflects the QA performance, which can more accurately estimate the question value and select useful pairs for target-domain QA.

\noindent (2) Two QVE baselines (binary classifier and ranking baseline) can select some useful questions and achieve comparable performance with RoundTrip and LM Filtering. However, due to the lack of direct QA evaluation feedback, they underperform QVE (RL), which demonstrates the usefulness of the QA feedback during training QVE. 



\subsection{How many target QA pairs do we need?}
\label{sec:target_dev_size}
In Table \ref{tbl:overall_performance}, we showed that with $n$ ($n$=1,000) target annotated QA pairs and the selected high-quality synthetic QA pairs, we can finetune a better QA model on the target domain. In this section, we discuss the influence of $n$ on the target-domain QA performance. The results are shown in Figure \ref{fig:number_of_target_samples}, and interesting findings include:

(1) In general, the performance of all models improves as more target annotations are used.  This is intuitive as more annotated pairs can improve both QA and QG training. With a better QG model, the quality of the synthetic questions is improved, which could also lead to better QA models.   

(2) Our QVE model can often outperform the QG baseline and the filtering baselines. With an optimization objective considering the downstream QA performance, QVE can select more useful questions for improving target-domain QA.

(3) The improvement of our QVE compared with baselines is usually larger when more annotated QA pairs are available. 
This is because our QVE training (with RL) relies on the QA feedback based on the available annotated pairs. With more annotated pairs, the feedback can be more accurate, thus leading to a better QVE for selecting more useful synthetic questions.

(4) With 10,000 (around 15\% of the original training set) target annotations and the synthetic questions selected by QVE, we can achieve comparable performance with the fully-supervised baseline. This indicates that one can save more annotation budgets when building a target-domain QA model based on our QVE in practice.

\subsection{Experiments with Larger Models}
\label{sec:other_qg_qa_model}
The results presented in the previous sections are based on \texttt{BERT-base} and \texttt{BART-base}. In this section, we test whether our QVE can still be effective when working with larger models, and select \texttt{BERT-Large} and \texttt{BART-Large}  as QA and QG model respectively. When changing the QA (QG) model to its larger alternative, we keep the other one as the base model to better show the difference. We use NaturalQuestions (NQ) and HotpotQA as representative datasets, and show results on them (with 1,000 target annotations).  As shown in Table \ref{tbl:other_qg_model}, our QVE model can still help improve the performance for larger instantiations of QG/QA.

\defcitealias{dong19unified}{UniLM}
\defcitealias{LeeLJKH20}{Info-HCVAE}
\defcitealias{Devlin19Bert}{Devlin, 2019}
\defcitealias{liu2019roberta}{(Liu, 2019)}

\begin{table}[t]
\centering
\resizebox{\linewidth}{!}{%
\begin{tabular}{l|l|cccc}
\hline
  \multirow{2}{*}{Setups}
  & \multirow{2}{*}{Methods}
  & \multicolumn{2}{c}{NQ} & \multicolumn{2}{c}{HotpotQA}  \\ \cline{3-6} 
  & & EM & F1 & EM & F1 \\ \hline

 \multirow{6}{*}{\begin{tabular}[c]{@{}l@{}}
QA:Large Model\\ 
QG:Base Model\end{tabular}}
& Source Only  &50.7	&65.0	&46.2	&64.0 \\ \cline{2-6} 
& + Target Annot.  &58.7 &72.1	&54.3	&72.2		 \\\cline{2-6} 
& + QG Baseline  &61.6 &73.4	&55.5	&72.5		 \\\cline{2-6} 
& + Roundtrip  &59.8 &71.9	&55.9	&72.8		 \\\cline{2-6} 
& + LM Filtering  &60.6 &72.5	&55.7	&72.7		 \\\cline{2-6} 
& \textbf{+ QVE (RL)}  & \textbf{62.4} &\textbf{74.5}	&\textbf{56.3}	&\textbf{73.4}	\\
 \hline
 \multirow{6}{*}{\begin{tabular}[c]{@{}l@{}}
QA:Base Model\\ 
QG:Large Model\end{tabular}}
& Source Only  &45.2	&59.1	&43.3	&60.3 \\ \cline{2-6}
& + Target Anno.  & 54.2 &68.2	&51.7	&69.2		 \\\cline{2-6} 
& + QG Baseline   &61.0	&72.8	&53.2	&70.9	 \\\cline{2-6} 
& + Roundtrip  &59.9 &71.7	&54.1	&71.1		 \\\cline{2-6} 
& + LM Filtering  &60.6 &72.2	&54.2	&71.2		 \\\cline{2-6} 
& \textbf{+ QVE (RL)}  &\textbf{62.1}	&\textbf{73.8}	&\textbf{55.2}	&\textbf{72.0}	\\
 \hline
\end{tabular}
}
\caption{Results on larger capacity QG and QA models.}
\label{tbl:other_qg_model}
\end{table}


\begin{table*}[!t]
\scriptsize
\resizebox{\linewidth}{!}{%
\begin{tabular}{lll|cc|ccc}
\hline
\multirow{2}{*}{\begin{tabular}[c]{@{}l@{}}Question ID \\ in the dataset\end{tabular}} & \multirow{2}{*}{Context} & \multirow{2}{*}{Question} & \multicolumn{2}{c|}{Human Labels} & \multicolumn{3}{c}{Selected by models?} \\ \cline{4-8} 
 &  &  & Matched & \begin{tabular}[c]{@{}c@{}}Non-\\ Trivial\end{tabular} & Roundtrip & LM & \begin{tabular}[c]{@{}c@{}}QVE \\ (Ours)\end{tabular} \\ \hline
\begin{tabular}[c]{@{}l@{}}NewsQA\\ \\ ./cnn/stories/\\ 6573f73a89\\ 7ec00e2c03\\ 7f959d832d\\ 04aa1a5ab3\\ .story\#1\end{tabular} & \begin{tabular}[c]{@{}l@{}}...Police arrested alleged ringleaders \\ Deborah Turbiville and her husband, \\ Charlie, as part of a two-year investigation, \\ the affiliate reported. Turbiville called\\ herself the "Heidi Fleiss of Houston," \\ referring to a woman who was dubbed \\ the \textless{}ANS\textgreater "Hollywood Madam" \\ \textless{}ANS\textgreater for providing call girls to \\ famous and wealthy clients, police said.\end{tabular} & \begin{tabular}[c]{@{}l@{}}What was the \\ nickname given \\ to the woman \\ who allegedly \\ provided call \\ girls for \\ prostitution?\end{tabular} & 1 & 1 & 0 & 0 & 1 \\ \hline
\begin{tabular}[c]{@{}l@{}}NQ\\ \\ aeee2c92\\ 647541da\\ 963bdb80\\ c5efc375\end{tabular} & \begin{tabular}[c]{@{}l@{}}...I 'm singing ' Pretending someone else \\ can come and save me from myself ' during \\it because it 's supposed to feel like an \\ apology letter , as though I 'm moving on\\ but I want people to remember the goodthings\\ and not the bad things. \textless{}ANS\textgreater A lot of the song \\is about humility \textless{}ANS\textgreater . ''...\end{tabular} & \begin{tabular}[c]{@{}l@{}}What is a lot \\ of the song \\ about?\end{tabular} & 1 & 0 & 1 & 1 & 0 \\ \hline
\end{tabular}
}
\caption{Two synthetic questions labeled by human and different question selection models. }
\label{apdx_tbl:case_study}
\end{table*}

\subsection{Human Study: Why can QVE help QA?}
In this section, we aim to gain a better understanding of why QVE helps QA and verify that QVE selects more semantically matched and non-trivial questions, thus benefiting downstream QA.

Since automatic metrics cannot often reflect the actual quality of the question selections, we sample 50 generated examples from each target-domain dataset (200 in total), and ask three human annotators to label whether a generated QA pair is semantically matched (i.e., can be selected to train QA) and (if yes) whether it asks about a simple fact. To lower the annotation bias in determining whether a generated question asks about a simple fact or not, we provide the ground-truth question (the question in the original dataset created by humans) as a reference. If the generated question is simpler than the ground truth, then it would be marked as ``trivial''; otherwise, it is a ``non-trivial'' one. Three annotators work independently and we adopt the majority vote for deciding the final labels of a generated QA pair (if disagreement appears).

We calculate the precision, recall and F1 between predictions\footnote{We treat it as a binary classification problem here: if a question is selected, the prediction is 1; 0 otherwise.} by each filtering method and human labels (for both ``semantically matched'' and ``non-trivial''). 
As shown in Table \ref{tbl:human_study}, though three methods obtain a similar precision on all sampled questions, our method has a better recall, especially on the ``non-trivial'' questions. This means that our method can select more semantically matched and non-trivial questions, which explains why it leads to better QA performance. We also show some real cases in Figure  \ref{fig:intro} and Table \ref{apdx_tbl:case_study} to further illustrate this point. For example, our QVE selects \textit{``What was the nickname given to the woman who allegedly provided call girls for prostitution?''} while the baselines do not pick this semantically matched and non-trivial question. For another example, \textit{``Who is the founder of CNN''}, both baselines select it while our QVE filters it out since such a simple question would probably not help further improve QA.
\begin{table}[!t]
\centering
\resizebox{\linewidth}{!}{%
    \begin{tabular}{l|c|c|c|c|c|c}
    \hline
     \multirow{2}{*}{Methods} & \multicolumn{3}{c|}{Semantically-Matched} & \multicolumn{3}{c}{Non-trivial} \\ \cline{2-7}
     & P & R & F1 & P & R & F1 \\ \hline
    RoundTrip & 87.9 &60.0  &71.2  &82.6  & 47.5 &60.3 \\ \hline
    LM Filtering &85.7  &64.6  &73.6  &78.9  &51.7  &62.5 \\ \hline
    \textbf{QVE(RL)} & \textbf{88.2} & \textbf{70.0} & \textbf{78.0}  &\textbf{83.3}  &\textbf{59.3}  &\textbf{69.3} \\ \hline
    \end{tabular}%
}
\caption{Agreement with question selection by humans.}
\label{tbl:human_study}
\end{table}

\section{Conclusion}
We propose a question value estimator to estimate the usefulness of synthetic questions and select useful ones for improving target-domain QA training. We optimize QVE with the target-domain QA performance gain after adding the selected questions into training. Our comprehensive experiments demonstrate the superiority of QVE compared with other question selection methods. Additionally, using the synthetic questions selected by QVE and only around 15\% of the human annotated data on each target domain, we can achieve comparable performance to the fully-supervised baselines.  \nop{With the synthetic questions selected by QVE\nop{and only 15\% of the full training set \st{annotated data}}, we can achieve comparable performance to the fully-supervised baselines while using only 15\% of their training set. }

\section*{Acknowledgement}
The authors would thank all the anonymous reviewers and the entire OSU and GMU NLP Group. This research was sponsored in part by NSF IIS-1815674, NSF CAREER \#1942980, NSF OAC-2112606, and Ohio Supercomputer Center \cite{OhioSupercomputerCenter1987}. The views and conclusions contained herein are those of the authors and should not be interpreted as representing the official policies,
either expressed or implied, of the U.S.Government. The U.S. Government is authorized to reproduce and distribute reprints for Government purposes notwithstanding any copyright notice herein.


\begin{thebibliography}{45}
\expandafter\ifx\csname natexlab\endcsname\relax\def\natexlab#1{#1}\fi

\bibitem[{Alberti et~al.(2019)Alberti, Andor, Pitler, Devlin, and
  Collins}]{Alberti19Synthetic}
Chris Alberti, Daniel Andor, Emily Pitler, Jacob Devlin, and Michael Collins.
  2019.
\newblock \href {https://doi.org/10.18653/v1/p19-1620} {Synthetic {QA} corpora
  generation with roundtrip consistency}.
\newblock In \emph{{ACL}'19}, pages 6168--6173. Association for Computational
  Linguistics.

\bibitem[{Cao et~al.(2020)Cao, Fang, Yu, and Zhou}]{cao2020unsupervised}
Yu~Cao, Meng Fang, Baosheng Yu, and Joey~Tianyi Zhou. 2020.
\newblock Unsupervised domain adaptation on reading comprehension.
\newblock In \emph{AAAI'20}.

\bibitem[{Chen et~al.(2016)Chen, Xu, Zhang, and Guestrin}]{corr/ChenXZG16}
Tianqi Chen, Bing Xu, Chiyuan Zhang, and Carlos Guestrin. 2016.
\newblock \href {http://arxiv.org/abs/1604.06174} {Training deep nets with
  sublinear memory cost}.
\newblock \emph{CoRR}, abs/1604.06174.

\bibitem[{Chen et~al.(2020)Chen, Sultan, and Castelli}]{chen20improved}
Yanda Chen, Md.~Arafat Sultan, and Vittorio Castelli. 2020.
\newblock \href {http://arxiv.org/abs/2010.12776} {Improved synthetic training
  for reading comprehension}.
\newblock \emph{CoRR}, abs/2010.12776.

\bibitem[{Chung et~al.(2018)Chung, Lee, and Glass}]{chung2018supervised}
Yu-An Chung, Hung-Yi Lee, and James Glass. 2018.
\newblock Supervised and unsupervised transfer learning for question answering.
\newblock In \emph{NAACL-HLT'18}, pages 1585--1594.

\bibitem[{Devlin et~al.(2019)Devlin, Chang, Lee, and Toutanova}]{Devlin19Bert}
Jacob Devlin, Ming{-}Wei Chang, Kenton Lee, and Kristina Toutanova. 2019.
\newblock \href {https://doi.org/10.18653/v1/n19-1423} {{BERT:} pre-training of
  deep bidirectional transformers for language understanding}.
\newblock In \emph{{NAACL-HLT}'19}, pages 4171--4186. Association for
  Computational Linguistics.

\bibitem[{Dhingra et~al.(2018)Dhingra, Danish, and
  Rajagopal}]{dhingra2018simple}
Bhuwan Dhingra, Danish Danish, and Dheeraj Rajagopal. 2018.
\newblock Simple and effective semi-supervised question answering.
\newblock In \emph{NAACL'18}, pages 582--587.

\bibitem[{Dong et~al.(2019)Dong, Yang, Wang, Wei, Liu, Wang, Gao, Zhou, and
  Hon}]{li2019Unified}
Li~Dong, Nan Yang, Wenhui Wang, Furu Wei, Xiaodong Liu, Yu~Wang, Jianfeng Gao,
  Ming Zhou, and Hsiao{-}Wuen Hon. 2019.
\newblock \href
  {http://papers.nips.cc/paper/9464-unified-language-model-pre-training-for-natural-language-understanding-and-generation}
  {Unified language model pre-training for natural language understanding and
  generation}.
\newblock In \emph{NeurIPS'19}, pages 13042--13054.

\bibitem[{Du and Cardie(2018)}]{du2018harvesting}
Xinya Du and Claire Cardie. 2018.
\newblock Harvesting paragraph-level question-answer pairs from wikipedia.
\newblock In \emph{ACL'18}, pages 1907--1917.

\bibitem[{Du et~al.(2017)Du, Shao, and Cardie}]{du2017learning}
Xinya Du, Junru Shao, and Claire Cardie. 2017.
\newblock \href {https://www.aclweb.org/anthology/P17-1123.pdf} {Learning to
  ask: Neural question generation for reading comprehension}.
\newblock In \emph{ACL'17}, pages 1342--1352.

\bibitem[{Fabbri et~al.(2020)Fabbri, Ng, Wang, Nallapati, and
  Xiang}]{fabbri2020template}
Alexander~Richard Fabbri, Patrick Ng, Zhiguo Wang, Ramesh Nallapati, and Bing
  Xiang. 2020.
\newblock Template-based question generation from retrieved sentences for
  improved unsupervised question answering.
\newblock In \emph{ACL'20}, pages 4508--4513.

\bibitem[{Fisch et~al.(2019)Fisch, Talmor, Jia, Seo, Choi, and
  Chen}]{FischTJSCC19}
Adam Fisch, Alon Talmor, Robin Jia, Minjoon Seo, Eunsol Choi, and Danqi Chen.
  2019.
\newblock \href {https://doi.org/10.18653/v1/D19-5801} {{MRQA} 2019 shared
  task: Evaluating generalization in reading comprehension}.
\newblock In \emph{MRQA@EMNLP'19}, pages 1--13. Association for Computational
  Linguistics.

\bibitem[{Ghorbani and Zou(2019)}]{GhorbaniZ19}
Amirata Ghorbani and James~Y. Zou. 2019.
\newblock \href {http://proceedings.mlr.press/v97/ghorbani19c.html} {Data
  shapley: Equitable valuation of data for machine learning}.
\newblock In \emph{{ICML}'19}, volume~97 of \emph{Proceedings of Machine
  Learning Research}, pages 2242--2251. {PMLR}.

\bibitem[{Golub et~al.(2017)Golub, Huang, He, and Deng}]{GolubHHD17}
David Golub, Po{-}Sen Huang, Xiaodong He, and Li~Deng. 2017.
\newblock \href {https://doi.org/10.18653/v1/d17-1087} {Two-stage synthesis
  networks for transfer learning in machine comprehension}.
\newblock In \emph{{EMNLP}'17}, pages 835--844. Association for Computational
  Linguistics.

\bibitem[{Hazen et~al.(2019)Hazen, Dhuliawala, and Boies}]{hazen2019towards}
Timothy~J Hazen, Shehzaad Dhuliawala, and Daniel Boies. 2019.
\newblock Towards domain adaptation from limited data for question answering
  using deep neural networks.
\newblock \emph{arXiv preprint arXiv:1911.02655}.

\bibitem[{Jia et~al.(2019)Jia, Dao, Wang, Hubis, Hynes, G{\"{u}}rel, Li, Zhang,
  Song, and Spanos}]{JiaDWHHGLZSS19}
Ruoxi Jia, David Dao, Boxin Wang, Frances~Ann Hubis, Nick Hynes, Nezihe~Merve
  G{\"{u}}rel, Bo~Li, Ce~Zhang, Dawn Song, and Costas~J. Spanos. 2019.
\newblock \href {http://proceedings.mlr.press/v89/jia19a.html} {Towards
  efficient data valuation based on the shapley value}.
\newblock In \emph{{AISTATS}'19}, volume~89 of \emph{Proceedings of Machine
  Learning Research}, pages 1167--1176. {PMLR}.

\bibitem[{Joshi et~al.(2017)Joshi, Choi, Weld, and Zettlemoyer}]{JoshiCWZ17}
Mandar Joshi, Eunsol Choi, Daniel~S. Weld, and Luke Zettlemoyer. 2017.
\newblock \href {https://doi.org/10.18653/v1/P17-1147} {Triviaqa: {A} large
  scale distantly supervised challenge dataset for reading comprehension}.
\newblock In \emph{{ACL}'17}, pages 1601--1611. Association for Computational
  Linguistics.

\bibitem[{Kwiatkowski et~al.(2019)Kwiatkowski, Palomaki, Redfield, Collins,
  Parikh, Alberti, Epstein, Polosukhin, Devlin, Lee, Toutanova, Jones, Kelcey,
  Chang, Dai, Uszkoreit, Le, and Petrov}]{KwiatkowskiPRCP19}
Tom Kwiatkowski, Jennimaria Palomaki, Olivia Redfield, Michael Collins,
  Ankur~P. Parikh, Chris Alberti, Danielle Epstein, Illia Polosukhin, Jacob
  Devlin, Kenton Lee, Kristina Toutanova, Llion Jones, Matthew Kelcey,
  Ming{-}Wei Chang, Andrew~M. Dai, Jakob Uszkoreit, Quoc Le, and Slav Petrov.
  2019.
\newblock \href {https://transacl.org/ojs/index.php/tacl/article/view/1455}
  {Natural questions: a benchmark for question answering research}.
\newblock \emph{Trans. Assoc. Comput. Linguistics}, 7:452--466.

\bibitem[{Lee et~al.(2020)Lee, Lee, Jeong, Kim, and Hwang}]{LeeLJKH20}
Dong~Bok Lee, Seanie Lee, Woo~Tae Jeong, Donghwan Kim, and Sung~Ju Hwang. 2020.
\newblock \href {https://doi.org/10.18653/v1/2020.acl-main.20} {Generating
  diverse and consistent {QA} pairs from contexts with information-maximizing
  hierarchical conditional vaes}.
\newblock In \emph{{ACL}'20}, pages 208--224. Association for Computational
  Linguistics.

\bibitem[{Lewis et~al.(2020)Lewis, Liu, Goyal, Ghazvininejad, Mohamed, Levy,
  Stoyanov, and Zettlemoyer}]{Lewis20BART}
Mike Lewis, Yinhan Liu, Naman Goyal, Marjan Ghazvininejad, Abdelrahman Mohamed,
  Omer Levy, Veselin Stoyanov, and Luke Zettlemoyer. 2020.
\newblock \href {https://www.aclweb.org/anthology/2020.acl-main.703/} {{BART:}
  denoising sequence-to-sequence pre-training for natural language generation,
  translation, and comprehension}.
\newblock In \emph{{ACL}'20}, pages 7871--7880. Association for Computational
  Linguistics.

\bibitem[{Lewis et~al.(2019)Lewis, Denoyer, and Riedel}]{LewisDR19}
Patrick S.~H. Lewis, Ludovic Denoyer, and Sebastian Riedel. 2019.
\newblock \href {https://doi.org/10.18653/v1/p19-1484} {Unsupervised question
  answering by cloze translation}.
\newblock In \emph{{ACL}'19}, pages 4896--4910. Association for Computational
  Linguistics.

\bibitem[{Li et~al.(2020)Li, Wang, Dong, Wei, and Xu}]{li2020harvesting}
Zhongli Li, Wenhui Wang, Li~Dong, Furu Wei, and Ke~Xu. 2020.
\newblock Harvesting and refining question-answer pairs for unsupervised qa.
\newblock In \emph{ACL'20}, pages 6719--6728.

\bibitem[{Liu et~al.(2020)Liu, Wei, Niu, Chen, and He}]{liu2020asking}
Bang Liu, Haojie Wei, Di~Niu, Haolan Chen, and Yancheng He. 2020.
\newblock Asking questions the human way: Scalable question-answer generation
  from text corpus.
\newblock In \emph{WWW'20}, pages 2032--2043.

\bibitem[{Liu et~al.(2019)Liu, Song, Zou, and Zhang}]{acl/LiuSZZ19}
Miaofeng Liu, Yan Song, Hongbin Zou, and Tong Zhang. 2019.
\newblock \href {https://doi.org/10.18653/v1/p19-1189} {Reinforced training
  data selection for domain adaptation}.
\newblock In \emph{{ACL}}, pages 1957--1968. Association for Computational
  Linguistics.

\bibitem[{Nema et~al.(2019)Nema, Mohankumar, Khapra, Srinivasan, and
  Ravindran}]{nema2019let}
Preksha Nema, Akash~Kumar Mohankumar, Mitesh~M Khapra, Balaji~Vasan Srinivasan,
  and Balaraman Ravindran. 2019.
\newblock \href {https://www.aclweb.org/anthology/D19-1326.pdf} {Let’s ask
  again: Refine network for automatic question generation}.
\newblock In \emph{EMNLP-IJCNLP'19}, pages 3305--3314.

\bibitem[{OSC(1987)}]{OhioSupercomputerCenter1987}
OSC. 1987.
\newblock \href {http://osc.edu/ark:/19495/f5s1ph73} {Ohio supercomputer
  center}.

\bibitem[{Puri et~al.(2020)Puri, Spring, Shoeybi, Patwary, and
  Catanzaro}]{PuriSSPC20Training}
Raul Puri, Ryan Spring, Mohammad Shoeybi, Mostofa Patwary, and Bryan Catanzaro.
  2020.
\newblock \href {https://www.aclweb.org/anthology/2020.emnlp-main.468/}
  {Training question answering models from synthetic data}.
\newblock In \emph{{EMNLP}'20}, pages 5811--5826. Association for Computational
  Linguistics.

\bibitem[{Qu et~al.(2019)Qu, Ji, Qiu, Yang, Min, Chen, Huang, and
  Croft}]{wsdm/QuJQYMCHC19}
Chen Qu, Feng Ji, Minghui Qiu, Liu Yang, Zhiyu Min, Haiqing Chen, Jun Huang,
  and W.~Bruce Croft. 2019.
\newblock \href {https://doi.org/10.1145/3289600.3290978} {Learning to
  selectively transfer: Reinforced transfer learning for deep text matching}.
\newblock In \emph{{WSDM}}, pages 699--707. {ACM}.

\bibitem[{Rajpurkar et~al.(2016)Rajpurkar, Zhang, Lopyrev, and
  Liang}]{RajpurkarZLL16}
Pranav Rajpurkar, Jian Zhang, Konstantin Lopyrev, and Percy Liang. 2016.
\newblock \href {https://doi.org/10.18653/v1/d16-1264} {Squad: 100, 000+
  questions for machine comprehension of text}.
\newblock In \emph{EMNLP'16}, pages 2383--2392. The Association for
  Computational Linguistics.

\bibitem[{Rennie et~al.(2020)Rennie, Marcheret, Mallinar, Nahamoo, and
  Goel}]{RennieMMNG20}
Steven~J. Rennie, Etienne Marcheret, Neil Mallinar, David Nahamoo, and Vaibhava
  Goel. 2020.
\newblock \href {https://www.aclweb.org/anthology/2020.emnlp-main.87/}
  {Unsupervised adaptation of question answering systems via generative
  self-training}.
\newblock In \emph{{EMNLP}'20}, pages 1148--1157. Association for Computational
  Linguistics.

\bibitem[{Ruder and Plank(2017)}]{emnlp/RuderP17}
Sebastian Ruder and Barbara Plank. 2017.
\newblock \href {https://doi.org/10.18653/v1/d17-1038} {Learning to select data
  for transfer learning with bayesian optimization}.
\newblock In \emph{{EMNLP}}, pages 372--382. Association for Computational
  Linguistics.

\bibitem[{Shakeri et~al.(2020)Shakeri, dos Santos, Zhu, Ng, Nan, Wang,
  Nallapati, and Xiang}]{ShakeriSZNNWNX20}
Siamak Shakeri, C{\'{\i}}cero~Nogueira dos Santos, Henghui Zhu, Patrick Ng,
  Feng Nan, Zhiguo Wang, Ramesh Nallapati, and Bing Xiang. 2020.
\newblock \href {https://www.aclweb.org/anthology/2020.emnlp-main.439/}
  {End-to-end synthetic data generation for domain adaptation of question
  answering systems}.
\newblock In \emph{{EMNLP}'20}, pages 5445--5460. Association for Computational
  Linguistics.

\bibitem[{Sun et~al.(2018)Sun, Liu, Lyu, He, Ma, and Wang}]{sun2018answer}
Xingwu Sun, Jing Liu, Yajuan Lyu, Wei He, Yanjun Ma, and Shi Wang. 2018.
\newblock \href {https://doi.org/10.18653/v1/d18-1427} {Answer-focused and
  position-aware neural question generation}.
\newblock In \emph{EMNLP'18}, pages 3930--3939. Association for Computational
  Linguistics.

\bibitem[{Trischler et~al.(2017)Trischler, Wang, Yuan, Harris, Sordoni,
  Bachman, and Suleman}]{TrischlerWYHSBS17}
Adam Trischler, Tong Wang, Xingdi Yuan, Justin Harris, Alessandro Sordoni,
  Philip Bachman, and Kaheer Suleman. 2017.
\newblock \href {https://doi.org/10.18653/v1/w17-2623} {Newsqa: {A} machine
  comprehension dataset}.
\newblock In \emph{Rep4NLP@ACL'17}, pages 191--200. Association for
  Computational Linguistics.

\bibitem[{Tuan et~al.(2020)Tuan, Shah, and Barzilay}]{tuan2020capturing}
Luu~Anh Tuan, Darsh~J. Shah, and Regina Barzilay. 2020.
\newblock \href {https://aaai.org/ojs/index.php/AAAI/article/view/6440}
  {Capturing greater context for question generation}.
\newblock In \emph{AAAI'20}, pages 9065--9072. {AAAI} Press.

\bibitem[{Wang et~al.(2019)Wang, Gan, Liu, Liu, Gao, and Wang}]{WangGLLGW19}
Huazheng Wang, Zhe Gan, Xiaodong Liu, Jingjing Liu, Jianfeng Gao, and Hongning
  Wang. 2019.
\newblock \href {https://doi.org/10.18653/v1/D19-1254} {Adversarial domain
  adaptation for machine reading comprehension}.
\newblock In \emph{{EMNLP-IJCNLP}'19}, pages 2510--2520. Association for
  Computational Linguistics.

\bibitem[{Wiese et~al.(2017)Wiese, Weissenborn, and Neves}]{wiese2017neural}
Georg Wiese, Dirk Weissenborn, and Mariana Neves. 2017.
\newblock Neural domain adaptation for biomedical question answering.
\newblock In \emph{CoNLL'17}, pages 281--289.

\bibitem[{Williams(1992)}]{Williams92}
Ronald~J. Williams. 1992.
\newblock \href {https://doi.org/10.1007/BF00992696} {Simple statistical
  gradient-following algorithms for connectionist reinforcement learning}.
\newblock \emph{Mach. Learn.}, 8:229--256.

\bibitem[{Wolf et~al.(2020)Wolf, Debut, Sanh, Chaumond, Delangue, Moi, Cistac,
  Rault, Louf, Funtowicz, Davison, Shleifer, von Platen, Ma, Jernite, Plu, Xu,
  Scao, Gugger, Drame, Lhoest, and Rush}]{wolf-etal-2020-transformers}
Thomas Wolf, Lysandre Debut, Victor Sanh, Julien Chaumond, Clement Delangue,
  Anthony Moi, Pierric Cistac, Tim Rault, Rémi Louf, Morgan Funtowicz, Joe
  Davison, Sam Shleifer, Patrick von Platen, Clara Ma, Yacine Jernite, Julien
  Plu, Canwen Xu, Teven~Le Scao, Sylvain Gugger, Mariama Drame, Quentin Lhoest,
  and Alexander~M. Rush. 2020.
\newblock \href {https://www.aclweb.org/anthology/2020.emnlp-demos.6}
  {Transformers: State-of-the-art natural language processing}.
\newblock In \emph{EMNLP'20}, pages 38--45, Online. Association for
  Computational Linguistics.

\bibitem[{Yang et~al.(2018)Yang, Qi, Zhang, Bengio, Cohen, Salakhutdinov, and
  Manning}]{Yang0ZBCSM18}
Zhilin Yang, Peng Qi, Saizheng Zhang, Yoshua Bengio, William~W. Cohen, Ruslan
  Salakhutdinov, and Christopher~D. Manning. 2018.
\newblock \href {https://doi.org/10.18653/v1/d18-1259} {Hotpotqa: {A} dataset
  for diverse, explainable multi-hop question answering}.
\newblock In \emph{EMNLP'18}, pages 2369--2380. Association for Computational
  Linguistics.

\bibitem[{Yoon et~al.(2020)Yoon, Arik, and Pfister}]{yoon2020data}
Jinsung Yoon, Sercan Arik, and Tomas Pfister. 2020.
\newblock \href {http://proceedings.mlr.press/v119/yoon20a/yoon20a.pdf} {Data
  valuation using reinforcement learning}.
\newblock In \emph{ICML'20}, pages 10842--10851. PMLR.

\bibitem[{Yue et~al.(2021)Yue, Kratzwald, and Feuerriegel}]{yue2021contrastive}
Zhenrui Yue, Bernhard Kratzwald, and Stefan Feuerriegel. 2021.
\newblock Contrastive domain adaptation for question answering using limited
  text corpora.
\newblock In \emph{EMNLP'21}.

\bibitem[{Zhang and Bansal(2019)}]{ZhangB19Addressing}
Shiyue Zhang and Mohit Bansal. 2019.
\newblock \href {https://doi.org/10.18653/v1/D19-1253} {Addressing semantic
  drift in question generation for semi-supervised question answering}.
\newblock In \emph{{EMNLP-IJCNLP}'19}, pages 2495--2509. Association for
  Computational Linguistics.

\bibitem[{Zhao et~al.(2018)Zhao, Ni, Ding, and Ke}]{zhao2018paragraph}
Yao Zhao, Xiaochuan Ni, Yuanyuan Ding, and Qifa Ke. 2018.
\newblock \href {https://doi.org/10.18653/v1/d18-1424} {Paragraph-level neural
  question generation with maxout pointer and gated self-attention networks}.
\newblock In \emph{EMNLP'18}, pages 3901--3910. Association for Computational
  Linguistics.

\bibitem[{Zhou et~al.(2017)Zhou, Yang, Wei, Tan, Bao, and
  Zhou}]{zhou2017neural}
Qingyu Zhou, Nan Yang, Furu Wei, Chuanqi Tan, Hangbo Bao, and Ming Zhou. 2017.
\newblock \href {https://doi.org/10.1007/978-3-319-73618-1\_56} {Neural
  question generation from text: A preliminary study}.
\newblock In \emph{National CCF Conference on Natural Language Processing and
  Chinese Computing}, pages 662--671. Springer.

\end{thebibliography}

\clearpage
\appendix
\setcounter{table}{0}
\setcounter{figure}{0}
\renewcommand{\thetable}{A\arabic{table}}
\renewcommand{\thefigure}{A\arabic{figure}}

\section{Details of Datasets}
\label{apdx:datasets}
Specifically, following \citet{ShakeriSZNNWNX20}, we use \textbf{SQuAD 1.1} \cite{RajpurkarZLL16}, a large reading comprehension dataset that consists of 100k questions on more than 500 articles from Wikipedia, as the \textit{source-domain} dataset. For the \textit{target-domain} datasets, we consider the following 4 datasets since they are commonly used and have sufficient contexts to train the models.

\noindent\textbf{NewsQA} \cite{TrischlerWYHSBS17} consists of questions and answers based on a set of over 10k news articles from CNN News. 

\noindent\textbf{Natural Questions (NQ)} \cite{KwiatkowskiPRCP19} contains questions extracted from Google user search queries and passages from Wikipedia.

\noindent\textbf{HotpotQA} \cite{Yang0ZBCSM18} is a multi-hop question answering dataset based on Wikipedia passages.

\noindent\textbf{TriviaQA}
\cite{JoshiCWZ17} includes QA pairs authored by trivia enthusiasts, as well as evidence documents independently gathered from Web search results and Wikipedia articles.

\section{Impact of Synthetic Dataset Size}
\begin{figure}[!t]
    \centering
    \includegraphics[width=0.95\linewidth]{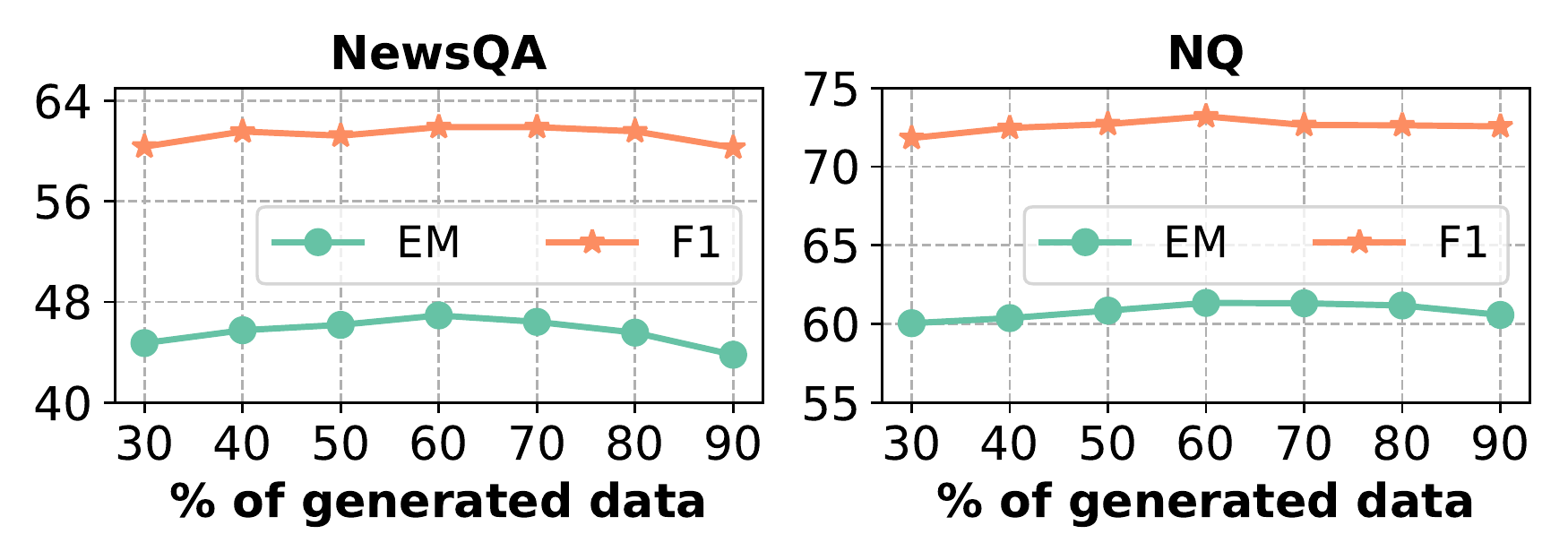}
    \caption{Impact of synthetic dataset size.}
    \label{fig:impact_dataset_size}
    \vspace{-5pt}
\end{figure}

In Figure \ref{fig:impact_dataset_size}, we show how the synthetic dataset size (i.e., the number of selected QA pairs) impacts the QA performance, based on our QVE (RL) filtering. As we expect, at the beginning, the target QA performance improves when more synthetic data is added to the training set. However, the performance reaches the peak at 60-70\% and then goes down. This is reasonable since adding less valuable QA pairs from the noisy synthetic data will hurt the QA model training.
We suggest 60\%-70\% (50K-70K QA pairs) for setting the synthetic data size in practical.

\label{sec:impact_syth_datasize}



\end{document}